\documentclass{jfm} 
\usepackage{hyperref, times, multicol}
\usepackage{xcolor,graphicx,caption,subcaption}
\usepackage{parskip,bm}
\usepackage{amsmath,amssymb,amsfonts,mathtools}

\begin{document}

\newtheorem{lemma}{Lemma}
\newtheorem{corollary}{Corollary}
\shorttitle{Gliding and Perching Through D-RL} 
\shortauthor{G. Novati et al.} 

\title{Deep-Reinforcement-Learning for \\ 
Gliding and Perching Bodies}

\author 
{
G. Novati \aff{1}
\and
L. Mahadevan \aff{2}
\and 
P. Koumoutsakos \aff{1} \corresp{\email{petros@ethz.ch}}
}
\affiliation 
{
\aff{1}
Computational Science and Engineering Laboratory, Clausiusstrasse~33, ETH~Z\"{u}rich, CH-8092, Switzerland
\aff{2}
John A. Paulson School of Engineering and Applied Sciences, Harvard University, Cambridge, MA 02138
}
  
\newcommand{\fix}{\marginpar{FIX}}
\newcommand{\new}{\marginpar{NEW}}

\maketitle
\begin{abstract} 
Controlled gliding is one of the most energetically efficient modes of transportation for natural and human powered fliers. Here we demonstrate that gliding and landing strategies with different optimality criteria can be identified  through deep reinforcement learning without explicit knowledge of the underlying physics. We combine a two dimensional model of a controlled elliptical body with deep reinforcement learning (D-RL) to achieve gliding with either minimum energy expenditure, or fastest time of arrival, at a predetermined location. In both cases the gliding trajectories are smooth, although energy/time optimal strategies are distinguished by small/high frequency actuations. We examine the effects of the ellipse's shape and weight on the optimal policies for controlled gliding. Surprisingly, we find that the model-free reinforcement learning leads to more robust gliding than model-based optimal control strategies with a modest additional computational cost. We also demonstrate that the gliders with D-RL can generalize their  strategies to reach the target location from previously unseen starting positions. The model-free character and robustness of D-RL suggests a promising framework for developing mechanical devices capable of exploiting complex flow environments. 
\end{abstract} 
\section{Introduction} 
Gliding is an intrinsically efficient motion that relies on the body shape to extract momentum from the air flow, while performing minimal mechanical work to control attitude. 
The sheer diversity of animal and plant species that have independently evolved the ability to glide is a testament to the efficiency and usefulness of this mode of transport. 
Well known examples include  birds that soar with thermal winds, fish that employ burst and coast swimming mechanisms and plant seeds, such as the samara, that spread by gliding.
Furthermore, arboreal animals that live in forest canopies often employ gliding to avoid earth-bound predators, forage across long distances, chase prey, and safely recover from falls.
Characteristic of gliding mammals is the membrane (patagium) that develops between legs and arms. When extended, the patagium transforms the entire body into a wing, 
allowing the mammal to stay airborne for extended periods of time~\cite{jackson2000}. Analogous body adaptations have developed in species of lizards~\cite{mori1994} and frogs~\cite{mccay2001}.

Most surprisingly, gliding has developed in animal species characterized by blunt bodies lacking specialized lift-generating appendages. The Chrysopelea genus of snakes have learned to launch themselves from trees, flatten and camber their bodies to form a concave cross-section, and perform sustained aerial undulations to generate enough lift to match the gliding performance of mammalian gliders~\citep{socha2002}. Wingless insects such as tropical arboreal ants~\citep{yanoviak2005} and bristletails~\citep{yanoviak2009} are able to glide when falling from the canopy in order to avoid the possibly flooded or otherwise hazardous forest understory. During descent these canopy-dwelling insects identify the target tree trunk using visual cues~\citep{yanoviak2006} and orient their horizontal trajectory appropriately.

Most bird species alternate active flapping with gliding, in order to reduce physical effort during long-range flight~\citep{rayner1985}. Similarly, gliding is an attractive solution to extend the range of micro air vehicles (MAVs).  MAV designs often rely on arrays of rotors (ie. quadcoptors) due to their simple structure and due to the existence of simplified models that capture the main aspects of the underlying fluid dynamics. The combination of these two features allows finding precise control techniques~\citep{gurdan2007,lupashin2010} to perform complex flight maneuvres~\citep{muller2011, mellinger2013}. However, the main drawback of rotor-propelled MAVs is their limited flight-times, which restricts real-world applications. Several solutions for extending the range of MAVs have been proposed, including techniques involving precise perching manouvres~\citep{thomas2016} and mimicking flying animals by designing a flier~\citep{abas2016} capable of gliding.

Here we study the ability of falling blunt-shaped bodies, lacking any specialized feature for generating lift, to learn gliding strategies through Reinforcement Learning \cite{bertsekas1995,kaelbling1996,sutton1998}. The goal of the RL agent is to control its descent towards a set target landing position and perching angle.
The agent is modeled by a simple dynamical system describing the passive planar gravity-driven descent of a cylindrical object in a quiescent fluid. The simplicity of the model is due to a parameterized model for the fluid forces which has been developed through simulations and experimental studies~\citep{wang2004, andersen2005a, andersen2005b}. Following the work of~\cite{paoletti2011}, we augment the original, passive dynamical system with active control. We identify optimal control policies through Reinforcement Learning, a semi-supervised learning framework, that has been employed successfully in a number of flow control problems~\citep{gazzola2014,gazzola2016,reddy2016,novati2017,colabrese2017}. We employ recent advances in coupling RL with deep neural networks~\cite{mnih2015,wang2016,novati2018}. These, so called Deep Reinforcement Learning algorithms have been shown in several problems to match and even surpass the performance of control policies obtained by classical approaches.

The paper is organised as follows: we describe the model of an active,  falling body in section~\ref{sec:model} and  frame the problems in terms of Reinforcement Learning in section~\ref{sec:rl}.  In section~\ref{sec:algo} we present a high-level description of the RL algorithm and describe the reward shaping combining the time/energy cost with kinematic constraints as described in section~\ref{sec:rew}. We explore the effects of the weight and shape of the agent's body on the optimal gliding strategies in section~\ref{sec:ret}. In sections~\ref{sec:vsoc} and~\ref{sec:comp} we compare the optimal RL policies by comparing them to other RL algorithms and the optimal control (OC) trajectories, e.g. ~\cite{paoletti2011}.  

\section{Model} \label{sec:model}
  We model the glider as an ellipse (see figure~\ref{fig:sketch}) with semi-axes $a \gg b$ and density $\rho_s$ in a quiescent incompressible fluid of density $\rho_f$. Under the assumption of planar motion, we can model the system with a set of ordinary differential equations (ODEs)~\citep{lamb1932}. The dimensionless form of the ODEs for the ellipse's translational and rotational degrees of freedom can written as \citep{andersen2005a,paoletti2011}:
\begin{figure} \centering
	\includegraphics[width=0.6\textwidth]{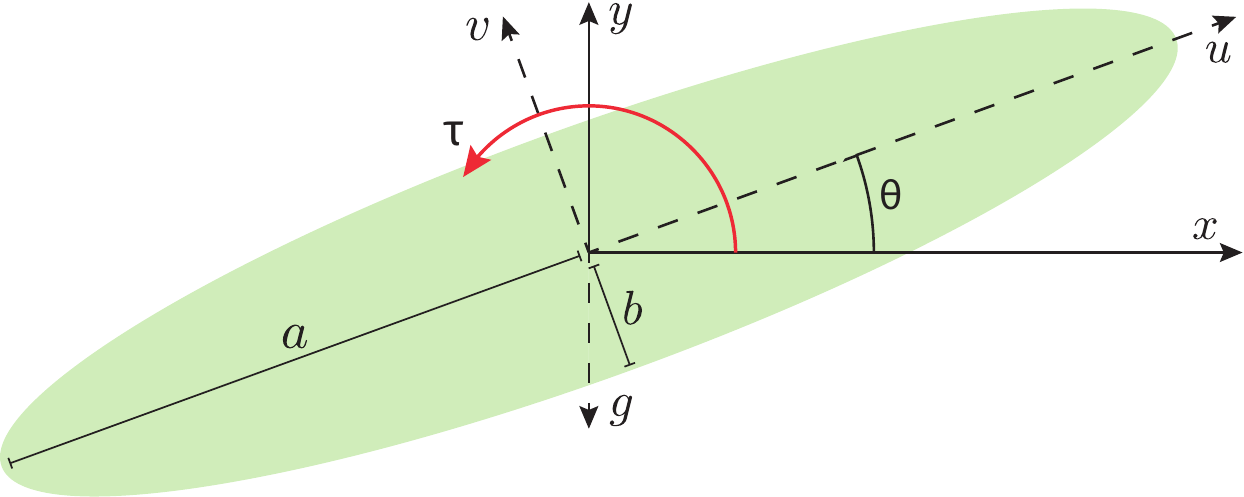} 
	\caption{Schematic of the system and degrees of freedom modeled by equations~\ref{eq:u} to~\ref{eq:a}. The position ($x$-$y$) of the center of mass is defined in a fixed frame of reference, $\theta$ is the angle between the $x$ axis and the major axis of the ellipse, and the velocity components $u$ and $v$ are defined in the moving and rotating frame of reference of the ellipse.}\label{fig:sketch}
\end{figure}
\begin{align}
(I{+}\beta^2)\dot{u} &= (I{+}1) vw - \Gamma v - \sin \theta  - F u \label{eq:u}\\
(I{+}1)\dot{v} &= - (I{+}\beta^2) uw + \Gamma u - \cos \theta  - F v\label{eq:v} \\
\frac{1}{4}\left[I(1{+}\beta^2) + \frac{1}{2}(1{-}\beta^2)^2\right] \dot{w} &= {-}(1{-}\beta^2) u v -M +\tau \label{eq:w}\\
\dot{x} &= u \cos \theta - v \sin \theta \label{eq:x}\\
\dot{y} &= u \sin \theta + v \cos \theta \label{eq:y}\\
\dot{\theta} &= w \label{eq:a}
\end{align}
Here $u$ and $v$ denote the projections of the velocity along the ellipse's semi-axes, $\theta$ is the angle between the major semi-axis and the horizontal direction, $w$ is the angular velocity, and $x$-$y$ is the position of the center of mass. Closure of the above system requires expressions for the forces and torques $F_u$, $F_v$, $M$, and the circulation $\Gamma$. Here, we approximate them in terms of a parametric model that has been developed through numerical and experimental studies by~\cite{wang2004,andersen2005a,andersen2005b}: 
\begin{align}
F &= \frac{1}{\pi} \left[A - B\frac{u^2 - v^2}{u^2 + v^2} \right] \sqrt{u^2 + v^2},\\
M &= 0.2 \left( \mu + \nu \| w\| \right) w, \\
\Gamma &= \frac{2}{\pi}\left[C_R w - C_T  \frac{uv}{\sqrt{u^2 + v^2}} \right] \label{eq:G}
\end{align}
Furthermore, the numerical constants are selected to be valid at intermediate Reynolds numbers $Re\approx o(10^3)$ based on the semi-major axes consistent with that of gliding ants~\cite{yanoviak2005}), with $A=1.4$, $B=1$, $\mu = \nu = 0.2$, $C_T = 1.2$, and $C_R = \pi$. Given the closure for the fluid forces and torques, the dynamics of the system are characterized by the non-dimensional parameters $\beta = b/a$, and $I  = \beta \rho^*$, where $\rho^*$ is the density ratio $\rho^* = \rho_s / \rho_f$.  

In active gliding, we assume that the gravity-driven descent can be modified by the agent by modulating the torque $\tau$ in equation~\ref{eq:w}. This torque could be achieved by deforming the body in order to move its center of mass, by deflecting the incoming flow, or by extending and rotating its limbs, as introduced  by~\cite{paoletti2011} as a minimal representation of the ability of gliding ants to guide their fall by rotating their hind legs~\cite{yanoviak2010}. This leads to a natural question: how should the active torque be varied in time for the ant to achieve a particular task such as landing and perching at a particular location with a particular orientation, subject to some constraints, e.g. optimizing time, minimizing power consumption, maximizing accuracy etc., a problem considered by ~\cite{paoletti2011} in an optimal control framework. Here we use an alternative approach inspired by how organisms might learn, that of reinforcement learning ~\cite{sutton1998}.

\section{Reinforcement Learning for landing and perching}\label{sec:rl}
The tasks of landing and perching are achieved by the falling body by employing a Reinforcement learning (RL) framework \cite{bertsekas1995,kaelbling1996,sutton1998} to identify their control actions. 

In the following we provide a brief overview of the RL framework in the context of flow control and outline the algorithms used in the present study.
Reinforcement Learning (RL) is a semi-supervised learning framework with a broad range of applications ranging from robotics~\cite{levine2016}, games~\cite{mnih2015,silver2016} and flow control \cite{gazzola2014}. In RL, the control actor (termed "agent")  interacts with its environment by sampling its states ($s$), performing  actions ($a$) and receiving  rewards ($r$). At each time step ($t$) the agent performs the action and the system is advanced  in time for $\Delta t$, before the agent can observe its new state $s_{t+1}$, receive a scalar reward $r_{t+1}$, and choose a new action $a_{t+1}$. The agent infers a policy $\pi (s,a)$ through its repeated interactions with the environment so as to maximize its long term rewards. The  optimal policy $\pi^* (s,a)$ is found by maximizing the expected utility:
\begin{equation}
J = \mathbb{E}_{\pi} \left[ \sum_{t=0}^{T} \gamma^{t} r_{t+1}\right]
\end{equation}
Once the optimal policy has been inferred the agent can interact autonomously with the environment. 

When the tasks of the agent satisfy the Markov property within  finite state and action spaces they are called finite Markov decision processes (MDP). Following \cite{sutton1998} the finite MDP is defined by the current state and action pair $(s,a)$ by the one step dynamics of the environment that are in turn described by  the probability $\mathcal{P}_{ss'}$ of any next possible state $s'$ and the expected value of the next rewards $\mathcal{R}_{ss'}$ defined as : 
\begin{equation} 
\mathcal{P}^{a}_{ss'} \,=\, Pr[s_{t+1}=s'|s_t=s, a_t=a], \;\;\;
\mathcal{R}^{a}_{ss'} \,=\, \mathbb{E}_{\pi} [r_{t+1}|s_t=s, a_t=a,s_{t+1}=s']
\end{equation}
The value function for each state $V^\pi (s)$ provides an estimate of future rewards given a certain policy $\pi$. For an MDP we define the state value function $V^{\pi}(s)$ as :
\begin{equation} 
V^\pi (s) = \mathbb{E}_{\pi} \left[ \sum_{k=0}^{\infty} \gamma^{k} r_{k+1} ~|~ s_0=s\right]
\end{equation}
where $\mathbb{E}_{\pi}$ denotes expected value for the agent when it performs the policy $\pi$ and $\gamma$ is a discount factor ($\gamma <1$) for future rewards. The action-value function $Q^\pi (s,a)$ satisfying the celebrated Bellman equation~\cite{Bellman1952} can be written as: 
\begin{equation}
Q^\pi(s,a) = 
\mathbb{E}_{\pi}\left[ \sum_{k=0}^{\infty} \gamma^{k} r_{k+1} ~|~ s_0=s,~ a_0=a\right]
\end{equation} 
The  state-value function is the expected action-value under the policy $\pi$: $ V^\pi(s) = \mathbb{E}_{a\sim\pi}\left[ Q^\pi(s,a) \right] $. 
We note that the $V^{\pi}(s), Q^\pi(s,a), \pi(s,a)$ may be described by tables that reflect discrete sets of states and actions. Such approaches have been used with some success in fluid mechanics applications \cite{gazzola2014,colabrese2017} but they often lead to poor training and are susceptible to noisy flow environments. In turn continuous approximations of such functions, such as those employed here in,  have been shown to lead to  robust and efficient learning policies~\cite{verma2018}.

We remark that the value functions and the Bellman equation are also inherent to Dynamic Programming (DP) \cite{bertsekas1995}. RL is  inherently linked to DP and it is often referred to as  approximate DP. The key difference between RL and dynamic programming is that RL does not require an accurate model of the environment and infers the optimal policy by a sampling process. 
Moreover RL can be applied to non MDPs. As such RL is in general more computationally intensive than DP and optimal control but at the same time can handle black-box problems and is robust to noisy and  stochastic environments. With the advancement of computational capabilities we believe that RL is becoming a valid complement to Optimal Control and other Machine Learning strategies \cite{duriez2017} for Fluid mechanics problems.


In the perching and landing context, we consider an agent, initially located at  $x(0)=y(0)=\theta(0)=0$, that has the objective of landing at a target location $x_G = 100$, $y_G = -50$ with perching angle $\theta_G = \frac{\pi}{4}$. By describing the trajectory with the model outlined in section~\ref{sec:model}, the state of the agent is completely defined at every time step by the  state vector $s := \{ x, y, \theta, u, v, w\}$. With a finite time interval $\Delta t = 0.5$, the agent is able to observe its state $s_t$ and, based on the state, samples a stochastic control policy $\pi$ to select an action $a_t \sim \pi (a | s_t)$. In the case of perching and landing an episode is terminated when at some terminal time $T$ the agent touches the ground $y_T = -50$. Because the gravitational force acting on the glider ensures that each trajectory will last a finite number of steps we can avoid the discount factor and set $\gamma = 1$. We consider continuous-valued controls defined by Gaussian policies which allows for fine-grained corrections(in contrast to usually employed  discretized controls~\cite{novati2017}). The action determines the constant control torque $\tau_t = \tanh (a_t)  \in \{-1, 1\} $ exerted by the agent between time $t$ and $t+\Delta t$.

\subsection{Off-policy actor-critic}\label{sec:algo}
We solve the RL problem with a novel off-policy actor-critic algorithm named {\tt Racer}~\citep{novati2018}. The algorithm relies on training a neural network (NN), defined by weights ${\tt w}$, to obtain a continuous approximation of  the policy $\pi^{\tt w}(a | s)$, the state value $V^{\tt w}(s)$ and the action value $Q^{\tt w}(s, a)$. The  network receives as input $s = \{ x, y, \theta, u, v, w\} \in \mathbb{R}^6$ and produces as output the set of parameters $\{m^{\tt w}, \sigma^{\tt w}, V^{\tt w}, l^{\tt w}\} \in \mathbb{R}^4$ that are further explained below.
The policy $\pi^{\tt w}(a | s)$ for each state is approximated with a Gaussian having a mean $m^{\tt w}(s)$ and a standard deviation $\sigma^{\tt w}(s)$:
\begin{equation}
\pi^{\tt w}(a | s) = \frac{1}{\sqrt{2 \pi \sigma^{\tt w}(s)}} ~\exp{\left[ -\frac{1}{2}\left(\frac{a-m^{\tt w}(s)}{\sigma^{\tt w}(s)}\right)^2 \right]} 
\end{equation}
The standard deviation is initially wide enough to adequately explore the dynamics of the system. In turn, we also suggest a continuous estimate of the state-action value function ($Q^\pi(s,a)$). Here, rather than having a specialized network, which includes in its input the action $a$, we propose computing the estimate $Q^{\tt w}(s, a)$ by combining the network's state value estimate $V^{\tt w}(s)$ with a quadratic term with vertex at the mean $m^{\tt w}(s)$ of the policy:
\begin{align}
Q^{\tt w}(s, a) &= V^{\tt w}(s)  -  \frac{1}{2} l^{\tt w}(s)^2 \left[a {-} m^{\tt w}(s)\right]^2 +  \mathop{\mathbb{E}}_{a'\sim\pi} \left[ \frac{1}{2} l^{\tt w}(s)^2 \left[a' {-} m^{\tt w}(s)\right]^2 \right]  \label{eq:qqpre}  \\
&= V^{\tt w}(s) - \frac{1}{2} l^{\tt w}(s)^2 \left[ \left[a {-} m^{\tt w}(s)\right]^2 - \left[\sigma^{\tt w}(s)\right]^2   \right]   \label{eq:qq}
\end{align}
This definition ensures that $ V^{\tt w}(s) = \mathbb{E}_{a\sim\pi}\left[ Q^{\tt w}(s,a) \right]$.
Here $l^{\tt w}(s)$ is an output of the network describing the rate at which the action value decreases for actions farther away from $m^{\tt w}(s)$. This parameterization relies on the assumption that for any given state $Q^{\tt w}(s, a)$ is maximal at the mean of the policy. 
 Since the dynamics of the system are  described by a small number of ordinary differential equations that can be solved at each instant to determine the state of the system, in contrast with the need to solve the full Navier-Stokes equations~\cite{novati2017}, we can use a continuous action space, and further use a multilayer perceptron ~\cite{sutton2000} rather than recurrent neural networks as policy approximators.

The learning process advances by iteratively sampling the dynamical system in order to assemble a set of training trajectories $\mathcal{B} = \{S_1, S_2, \dots\}$. A trajectory $S = \{ o_0, o_1, \dots, o_T\}$ is sequence of observations $o_t$. An observation is defined as the collection of all information available to the agent at time $t$: the state $s_t$, the, reward $r_t$, the current policy $\mu_t =\{ m_t, \sigma_t \}$ and the sampled action $a_t$. Here we made a distinction between the policy $\mu_t$ executed at time $t$ and $\pi^{\tt w}(a | s_t)$ because, when the data is used for training, the weights ${\tt w}$ of the NN might change, causing the current policy for state $s$ to change. 
For each new observation $o_t$ from the environment, a number $B$ of observations are sampled from the dataset $\mathcal{B}$.
Finally, the network weights are updated through back-propagation of the policy ($g_{\pi}$) and value function gradients ($g_\text{Q}$). 

The policy parameters $m^{\tt w}(s)$ and $\sigma^{\tt w}(s)$ are improved through the policy gradient estimator~\citep{Degris2012}: 
\begin{equation}
g_{\pi} = \sum_{t=1}^B  \frac{\pi^{\tt w}(a_t|s_t) }{\mu_t(a_t|s_t)} \left[\hat{Q}(s_t, a_t){-}V^{\tt w}(s_t) \right]  \nabla_{\tt w} \log \pi^{\tt w} (a_t | s_t) \label{eq:gradPi}
\end{equation} 
where $\hat{Q}(s_t, a_t)$ is an estimator  of the action value. A key insight from policy-gradient based algorithms is that the parameterized $Q^{\tt w}(s_t, a_t)$ cannot safely be used to approximate on-policy returns, due to its inaccuracy during training~\cite{sutton2000}. 
On the other hand, obtaining $Q^\pi(s_t, a_t)$ through Monte Carlo sampling is often computationally prohibitive. Hence, we approximate $\hat{Q}(s_t, a_t)$ with the Retrace algorithm~\cite{Munos2016}, which can we written recursively as:
\begin{align}
\hat{Q}(s_t, a_t) \approx \hat{Q}^\text{ret}(s_t, a_t) &= r_{t+1} +\gamma V^{\tt w}(s_{t+1}) \nonumber\\
& +\gamma \min \{1, \rho(s_t, a_t)\} \left[ \hat{Q}^\text{ret}(s_{t+1}, a_{t+1}) {-} Q^{\tt w}(s_{t+1}, a_{t+1})\right] \label{eq:retr}
\end{align}
The importance weight $\rho(s_t, a_t) = \pi^{\tt w}(a_t|s_t) / \mu_t(a_t|s_t)$, is the ratio of probabilities of sampling the action $a_t$ from state $s_t$ with the current policy $\pi^{\tt w}$ and with the old policy $\mu_t$. 

The state value $V^{\tt w}(s)$ and action value coefficient $l^{\tt w}(s)$ are trained with the importance-sampled gradient of the $L2$ distance from $\hat{Q}^\text{ret}$:
\begin{equation}
g_\text{Q} = \sum_{t=1}^B \frac{\pi^{\tt w}(a_t|s_t) }{\mu_t(a_t|s_t)} \left[ \hat{Q}^\text{ret}(s_t, a_t) - Q^{\tt w}(s_t, a_t) \right] \nabla_{\tt w} Q^{\tt w}(s_t, a_t) \label{eq:gradQ}
\end{equation}
Further implementation details of the algorithm can be found in~\cite{novati2018}.

\subsection{Reward formulation}\label{sec:rew}
We wish to identify energy-optimal and time-optimal control policies by varying the aspect and density ratios that define the system of ODEs. In the optimal control setting, boundary conditions, such as the initial and terminal positions of the ellipse, and constraints, such as bounds on the power or torque can be included directly in the problem formulation as employed in~\cite{paoletti2011}. 
In RL, boundary conditions can only be included in the reward formulation. The agent is discouraged from violating optimization constraints by introducing a condition for termination of a simulation, accompanied by negative terminal rewards.
For example, here we inform the agent about the landing target by composing the reward as:
\begin{equation}
r_{t} = -c_t + \| x_G - x_{t-1} \| - \| x_G - x_{t} \|
\end{equation} 
where $c_t$ is the optimal control cost function which can either be $\Delta t$ for learning time-optimal policies, or $\int_{t-1}^{t} \tau^2 \text{d}t =\tau^2 \Delta t $ for energy-optimal policies. The control cost $\tau^2$ is used as a proxy for the energy cost as in~\cite{paoletti2011}. Note that for a trajectory monotonically approaching $x_G$ the difference between the RL and optimal control cost functions $\sum_{t=0}^T r_t + c_t = \| x_G - x_{0} \| - \| x_G - x_{T} \|$. If the exact target location $x_G$ is reached at the terminal state, the discrepancy between the two formulations would be a constant baseline $ \| x_G - x_{0} \|$, which can be proved to not affect the policy~\cite{Ng1999}. Therefore, a RL agent that maximizes cumulative rewards also minimizes either the time or the energy cost.

The episodes are terminated if the ellipse touches the ground at $y_G=-50$. In order to allow the agent to explore diverse perching maneuvers, such as phugoid motions, the ground is recessed between $x=50$ and $x=100$ and is located at $y_G = -50 - 0.4 \min (x-50, 100-x)$.
For both time optimal and energy optimal optimizations, the terminal reward is given by:
\begin{equation}\label{eq:termr}
r_{T} = -c_T + K \left( e^{-(x_G - x_{T})^2} + e^{-10(\theta_G - \theta_{T})^2} \right)
\end{equation}
here $K{=}K_E {=} 20$ or $K{=}K_T {=} 50$ when training for energy- or time-optimal policies respectively. The second exponential term of Eq.~\ref{eq:termr} is added only if $95 < x_{T} < 105$, in order to prevent the policy from landing away from $x_G$ by relying on the perching angle bonus while minimizing time/energy costs. 

\section{Results}\label{sec:ret}

\begin{figure} \centering 
\begin{subfigure}{0.67\textwidth} 
	\includegraphics[width=\textwidth]{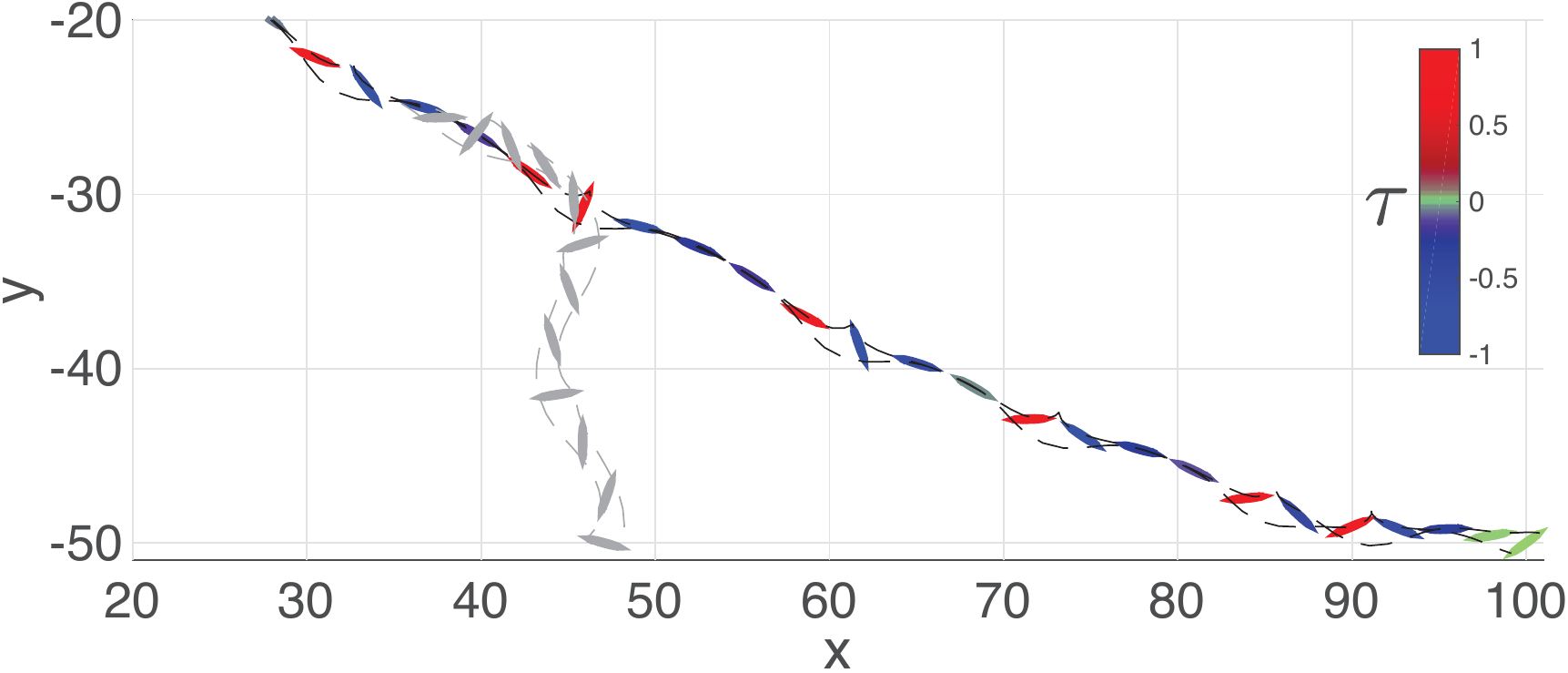}
	\subcaption{} \label{fig:bt} \end{subfigure} \hfill \begin{subfigure}{0.31\textwidth} \includegraphics[width=\textwidth]{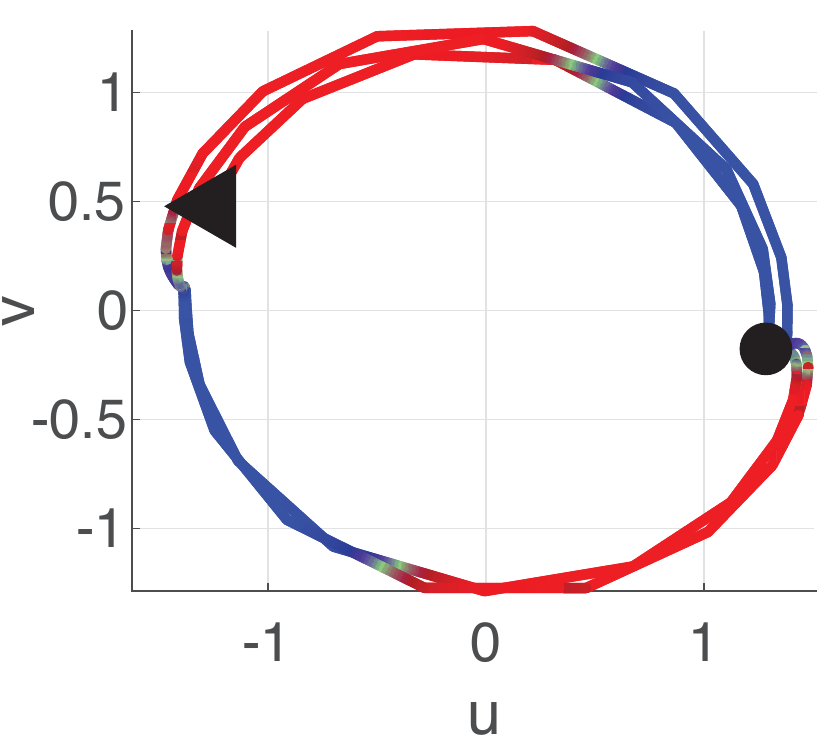}
	\subcaption{} \label{fig:bp} \end{subfigure} 
    \begin{subfigure}{0.67\textwidth} 
	\includegraphics[width=\textwidth]{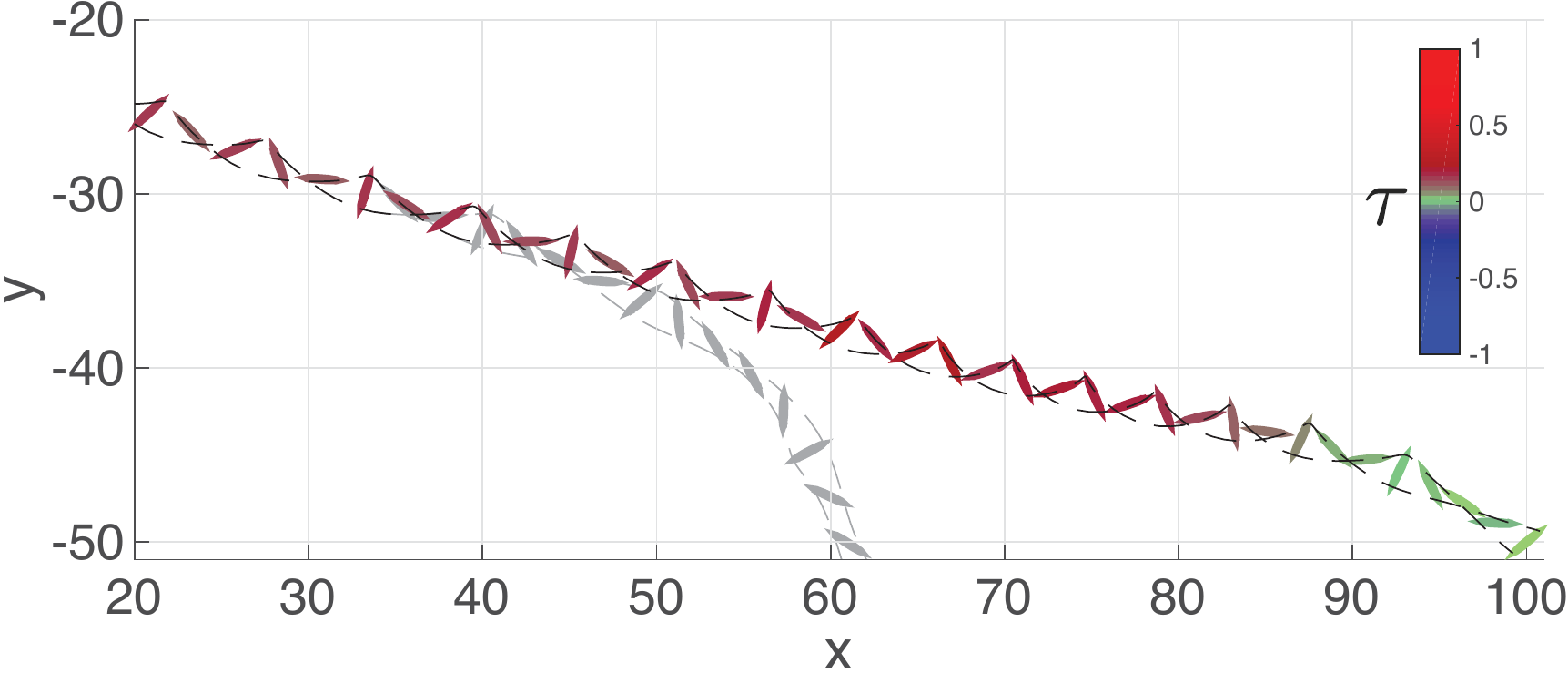}    
    \subcaption{} \label{fig:gt} \end{subfigure} \hfill\begin{subfigure}{0.31\textwidth} \includegraphics[width=\textwidth]{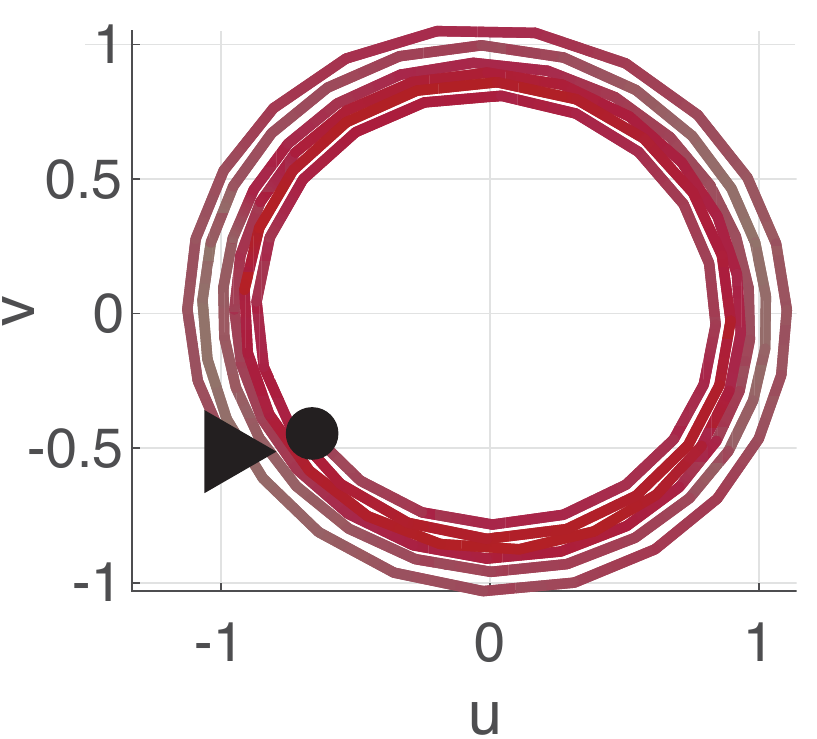}
	\subcaption{} \label{fig:gp} \end{subfigure} 
\caption{Visualization of the two prevailing locomotion patterns adopted by RL agents for the active gliding model described in Sec.~\ref{sec:model}. Trajectories on the $x$-$y$ plane for (\subref{fig:bt})~bounding ($\beta{=}0.1$, $\rho^*{=}100$) and (\subref{fig:gt})~tumbling flight ($\beta{=}0.1$, $\rho^*{=}200$). The glider's snaphots are colored to signal the value of the control torque, and the dashed black lines track the ellipse's vertices. The grayed-out trajectories illustrate the glider's passive descent when abruptly switching off active control. (\subref{fig:bp}, \subref{fig:gp}) Corresponding trajectories on the $u$-$v$ plane. For the sake of clarity, we omit the initial transient and final perching maneuverer. The trajectories are colored based on the control torque and their beginning and end are marked by a triangle and circle respectively.
 }\label{fig:patterns}
\end{figure}

We explore the gliding strategies of the RL agents that aim to minimize either time-to-target or energy expenditure, by varying the aspect ratio $\beta$ and density ratio $\rho^*$ of the falling ellipse. These two optimization objectives may be seen as corresponding to the biologic scenarios of foraging and escaping from predators. Figure~\ref{fig:patterns} shows the two prevailing flight patterns learned by the RL agent,which we refer to as `bounding' and `tumbling' flight ~\cite{paoletti2011}. The name `bounding' flight is due to an energy-saving flight strategy first analyzed by~\cite{rayner1977} and~\cite{lighthill1977} with simplified models of intermittently flapping fliers. 

In the present model, bounding flight is characterized by succeeding phases of gliding and tumbling. During gliding, the agent exerts negative torque to maintain a small angle of attack (represented by the blue snapshots of the glider in Fig.~\ref{fig:bt}), deflecting momentum from the air flow which slows down the descent. During the tumbling phase, the agent applies a rapid burst of positive torque (red snapshots of the glider in Fig.~\ref{fig:bt}) to generate lift and, after a rotation of $180^\circ$, recover into a gliding attitude. 

The trajectory on the $u$-$v$ plane (Fig.~\ref{fig:bp}) highlights that the sign of the control torque is correlated with whether $u$ and $v$ have the same sign. This behavior is consistent with the goal of maintaining upward lift. In fact, the vertical component of lift applied onto the falling ellipse is $\Gamma \dot{x}$, with $\dot{x}{>}0$ because the target position is to the right of the starting position. From Eq.\ref{eq:G} of our ODE-based model, the lift is positive if $u$ and $v$ have opposite signs or if $w$ is positive. Therefore, in order to create upward lift, the agent can either exert a positive $\tau$ to generate positive angular velocity, or, if $u$ and $v$ have opposite signs, exert a negative $\tau$ to reduce its angular velocity (Eq.\ref{eq:w}) and maintain the current orientation. 
The grayed-out trajectory shows what would happen during the gliding phase without active negative torque: the ellipse would increase its angle of attack, lose momentum and, eventually, fall vertically. 

Tumbling flight, visualized in figures~\ref{fig:gt} and~\ref{fig:gp}, is a much simpler pattern obtained by applying an almost constant torque that causes the ellipse to steadily rotate along its trajectory, thereby generating lift. The constant rotation is generally slowed down for the landing phase in order to descent and accurately perch at $\theta_G$.

\begin{figure} \centering 
\begin{subfigure}{0.325\textwidth} 
	\includegraphics[width=\textwidth, trim={3 1 20 14}, clip]{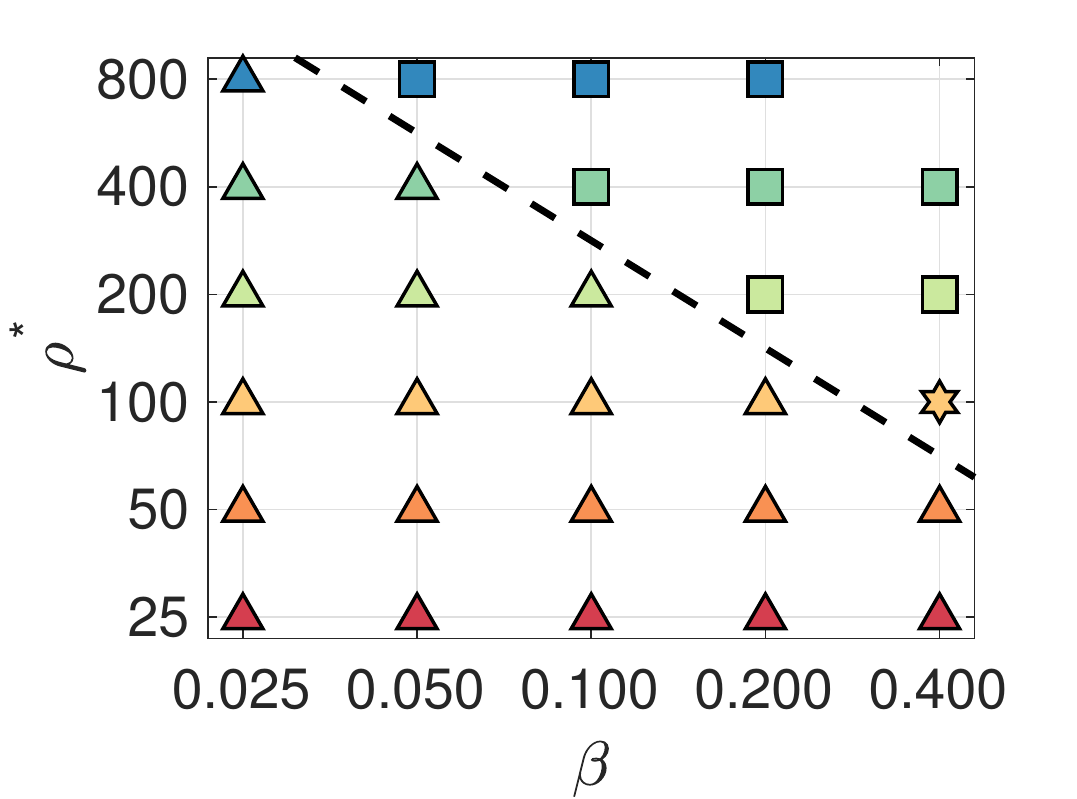}
	\subcaption{}\label{fig:expb} \end{subfigure} \hfill \begin{subfigure}{0.325\textwidth} 
	\includegraphics[width=\textwidth, trim={3 1 20 14}, clip]{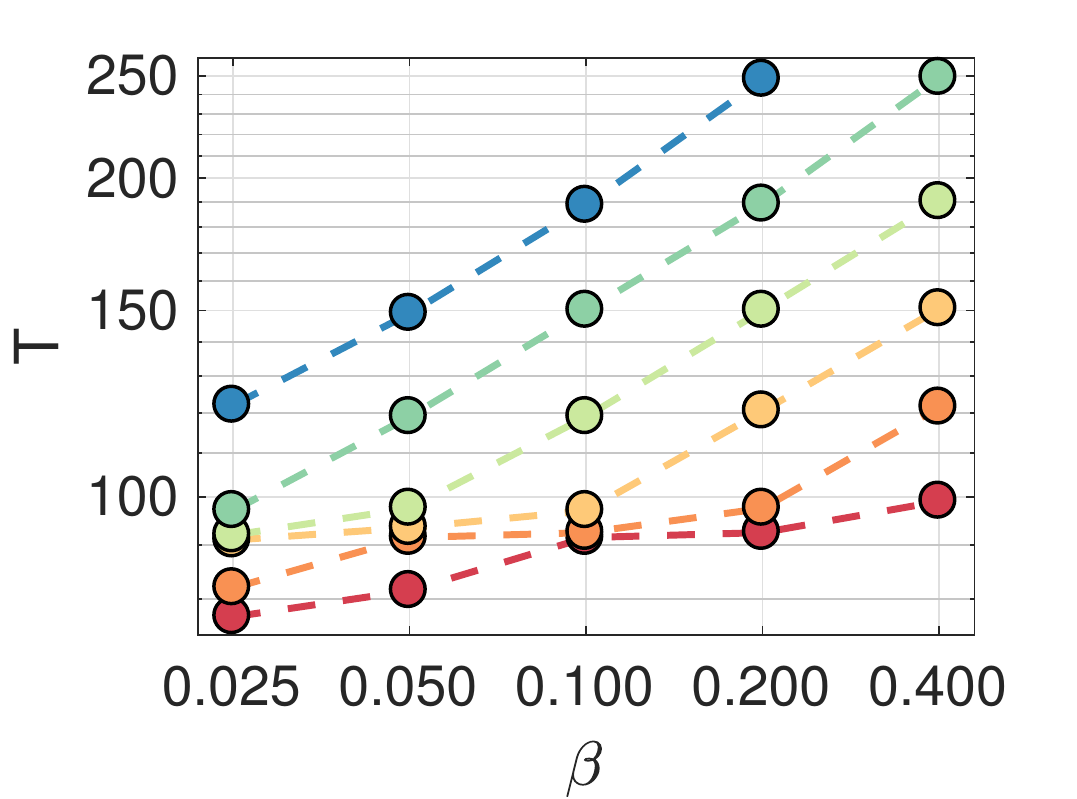}    
    \subcaption{}\label{fig:expt} \end{subfigure} \hfill \begin{subfigure}{0.325\textwidth} 
	\includegraphics[width=\textwidth, trim={3 1 20 14}, clip]{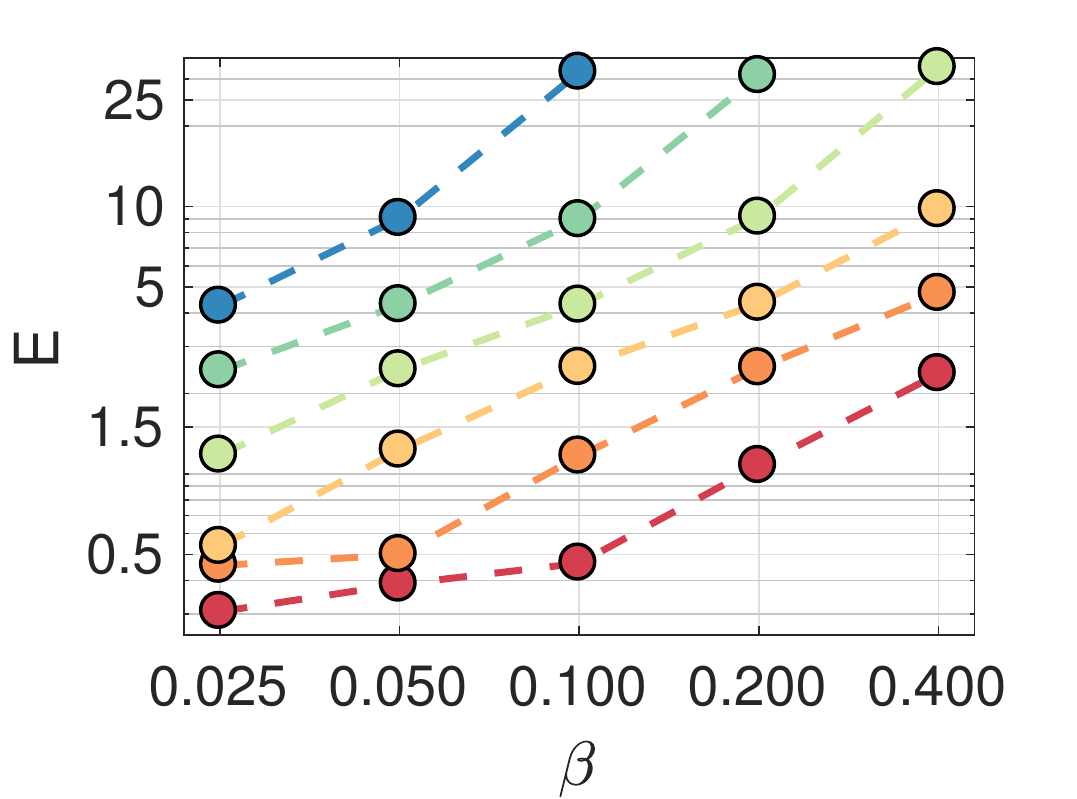}     
    \subcaption{}\label{fig:expe} \end{subfigure}
	\caption{Optimal solutions by sweeping the space of dimensionless parameters $\rho^*$ and $\beta$ of the ODE model outlined in Sec.~\ref{sec:model}. (\subref{fig:expb}) Flight pattern employed by time-optimal agents. Triangles refer to bounding flight and squares to tumbling. The policy for $\rho^*{=}100$ and $\beta{=}0.4$, marked by a star, alternated between the two patterns. The optimal (\subref{fig:expt}) time-cost and (\subref{fig:expe}) energy cost increase monotonically with both $\beta$ and $\rho^*$. The symbols are colored depending on the value of $\rho^*$: red for $\rho^*{=}25$, orange $\rho^*{=}50$, yellow $\rho^*{=}100$, lime $\rho^*{=}200$, green $\rho^*{=}400$, blue $\rho^*{=}800$. }\label{fig:phase}
\end{figure} 

In figure~\ref{fig:phase} we report the effect of the ellipse's shape and weight on the optimal strategies. The system of ODEs described in section~\ref{sec:model} is characterized by non-dimensional parameters $\beta$ and $I = \rho^* \beta$. Here we independently vary the density ratio $\rho^*$ and the aspect ratio $\beta$ in the range $[25, 800]\times[0.025, 0.4]$. For each set of dimensionless parameters we train a RL agent to find both the energy-optimal and time-optimal policies. The flight strategies employed by the RL agents can be clearly defined as either bounding flight or tumbling only in the time-optimal setting, while energy-optimal strategies tend to employ elements of both flight patterns. In figure~\ref{fig:expb}, time-optimal policies that employ bounding flight are marked by a triangle, while those that use tumbling flight are marked by a square. We find that lighter and elongated bodies employ bounding flight while heavy and thick bodies employ tumbling flight. Only one policy, obtained for $\rho^* = 100$, $\beta=0.4$ alternated between the two patterns and is marked by a star. These results indicate that a simple linear relation $\rho^* \beta = I \approx 30$ (outlined by a black dashed line in figure~\ref{fig:expb}) approximately describes the boundary between regions of the phase-space where one flight pattern is preferred over the other.
In figures~\ref{fig:expt} and~\ref{fig:expe} we report the optimal time costs and optimal energy costs for all the combinations of on-dimensional parameters.

\begin{figure} \centering
\begin{subfigure}{0.54\textwidth} 
	\includegraphics[width=\textwidth]{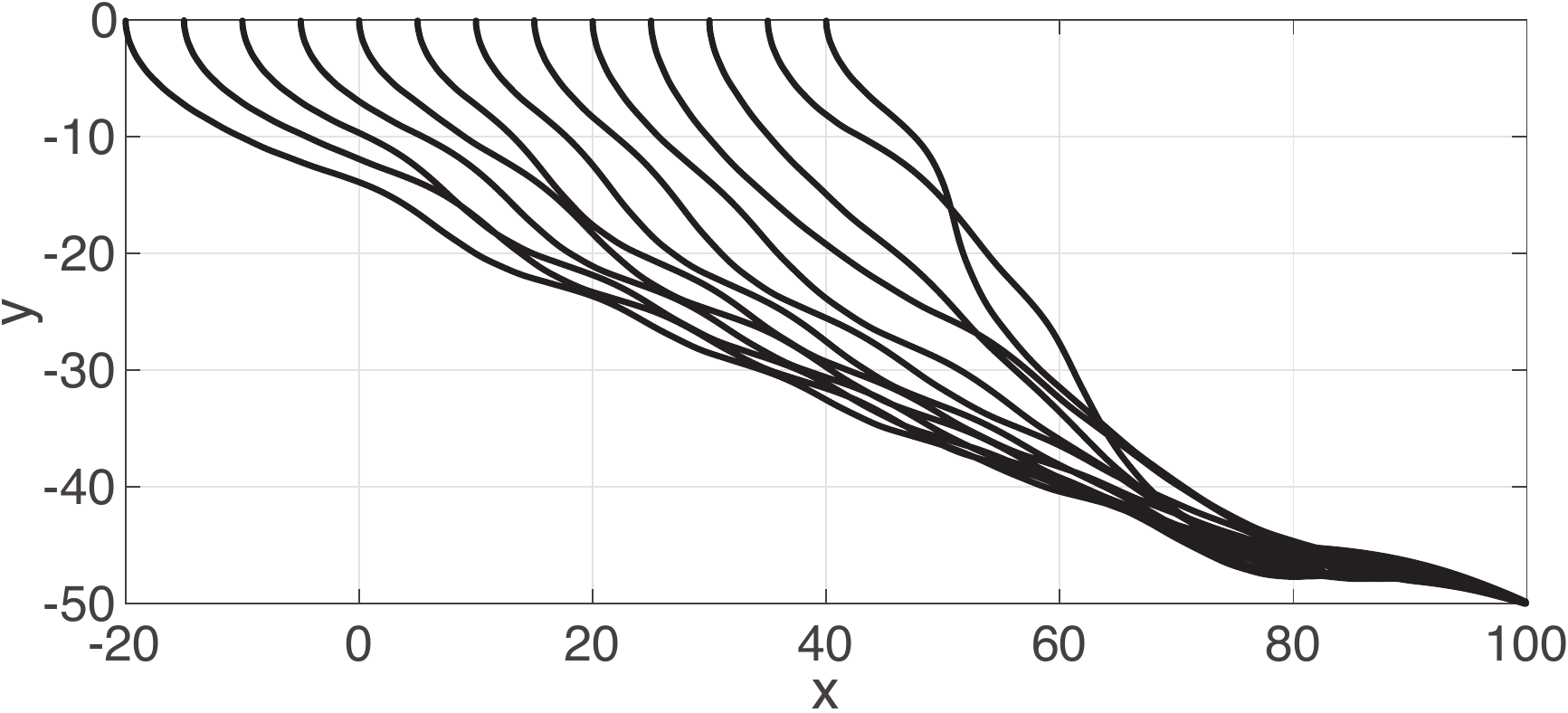}
	\subcaption{} \label{fig:noisex}\end{subfigure} \begin{subfigure}{0.45\textwidth} 
	\includegraphics[width=\textwidth]{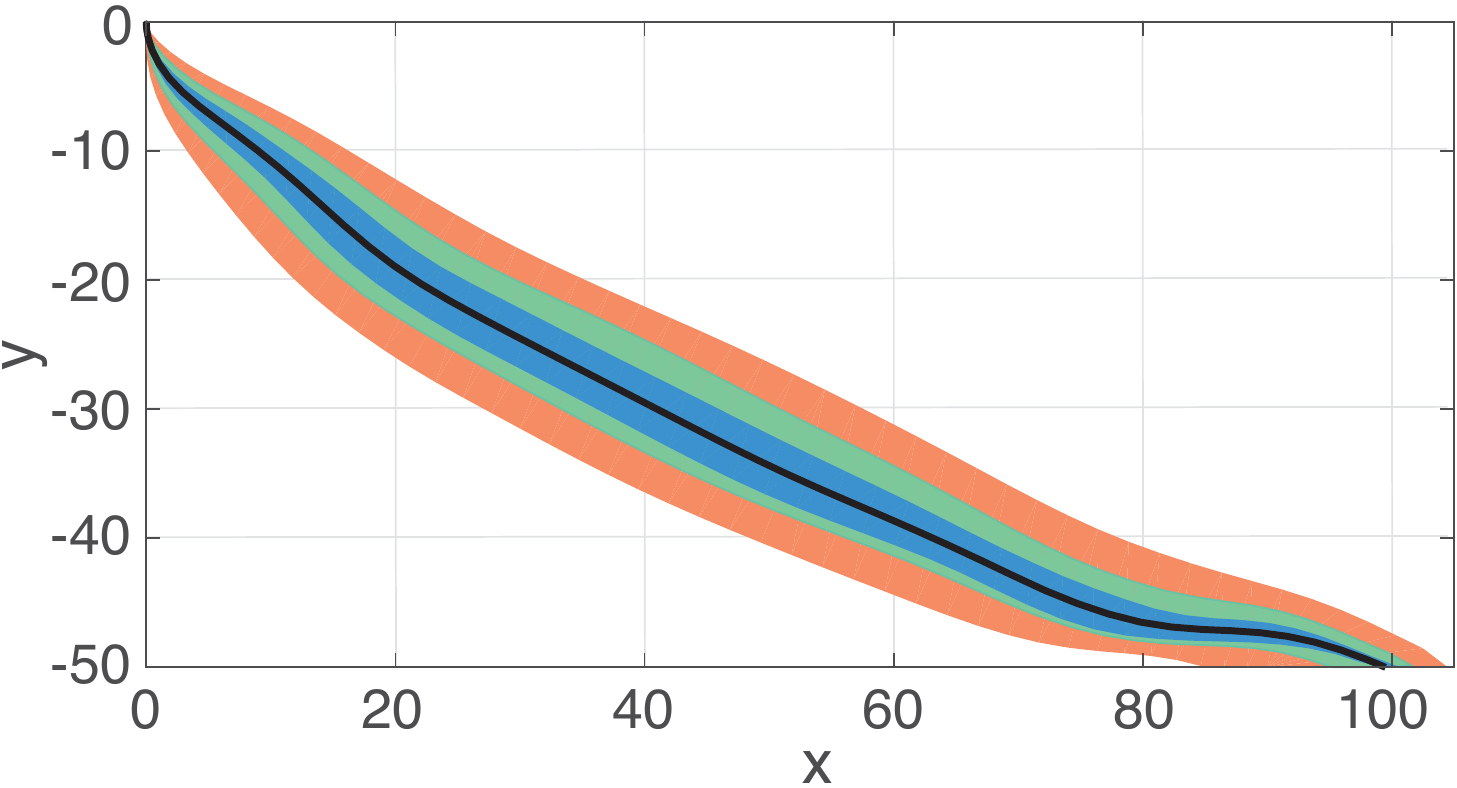}    
    \subcaption{} \label{fig:noisep} \end{subfigure}
	\caption{Robustness of the trained RL agent. 
    The agent is able to land in the neighborhood of the target $x_G$ despite (\subref{fig:noisex}) starting its trajectory from initial conditions not seen during training or (\subref{fig:noisep}) applying proportional noise to the parameters of the model outlined in Sec.~\ref{sec:model}. The color contours of Fig.~\subref{fig:noisep} represent the envelopes of $10^4$ trajectories for different values of standard deviation of the proportional log-normal noise. The blue contour corresponds to $\sigma_\xi = 0.1$, green to 0.2, orange to 0.4. These results are obtained with the time-optimal policy for $\beta{=}0.1$ and $\rho^*{=}200$.}\label{fig:noise}
\end{figure}

Once the RL training terminates, the agent obtains a set of opaque rules, parameterized by a neural network, to select actions. These rules are approximately-optimal only for the states encountered during training, but can also be applied to new conditions. In fact, we find that the policies obtained through RL are remarkably robust. In figure~\ref{fig:noisex} we apply the time-optimal policy for $\rho^*=200$ and $\beta=0.1$ to a new set of initial conditions along the $x$-coordinate. Despite the agent never having encountered these position during training, it can always manage to reach the perching target. Similarly, in figure~\ref{fig:noisep} we test the robustness with respect to changes to the parameters of the ODE model $\Psi = \{A, B, \mu, \nu, C_T, C_R\}$. At the beginning of a trajectory, we vary each parameter according to $\hat{\Psi}_i = \Psi_i \cdot \xi$ where $\xi$ is sampled from a log-normal distribution with mean 1 and standard deviation $\sigma_\xi$. The color contour of figure~\ref{fig:noisep} represent the envelopes of $10^4$ trajectories for $\sigma_\xi = 0.1$ (blue), 0.2 (green), and 0.4 (orange). Surprisingly, even when the parameters are substantially different from those of the original model, the RL agent always finds its bearing and manages to land in the neighborhood of the target position.

\section{Comparison with Optimal Control}\label{sec:vsoc}

\begin{figure} \centering 
    \begin{subfigure}{0.325\textwidth} 
	\includegraphics[width=\textwidth, trim={8 3 20 10}, clip]{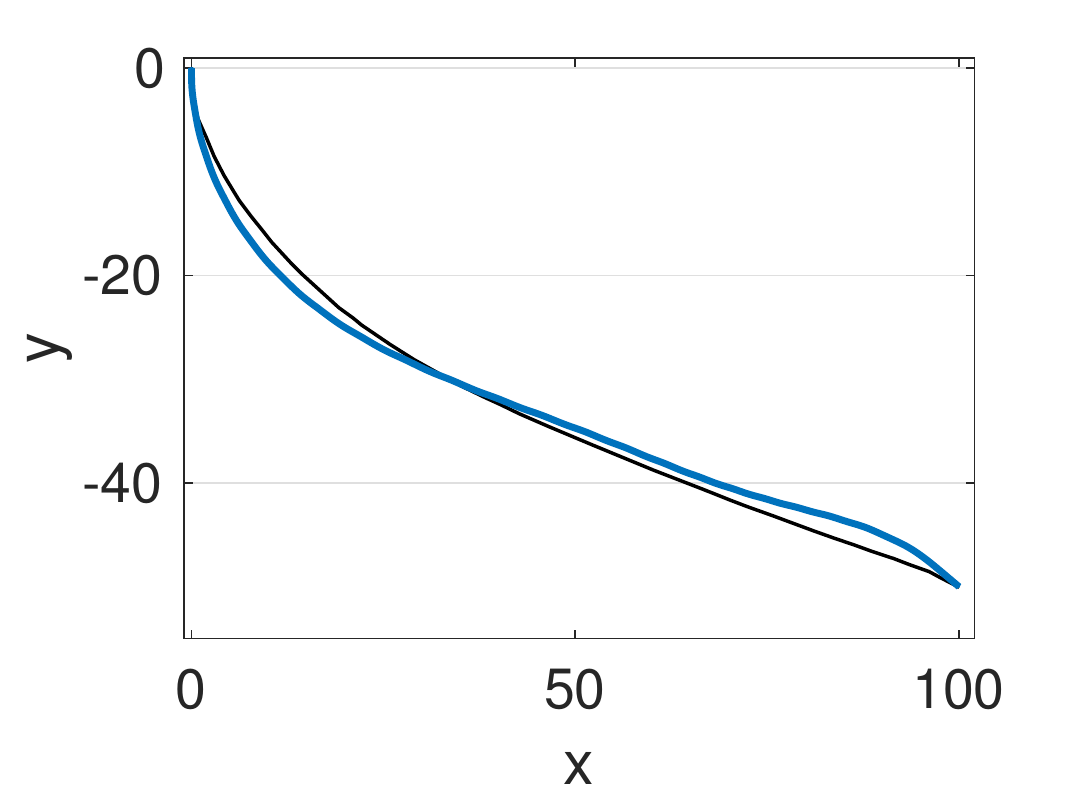}
	\subcaption{} \label{fig:enp}\end{subfigure} \hfill \begin{subfigure}{0.325\textwidth} 
	\includegraphics[width=\textwidth, trim={8 3 20 10}, clip]{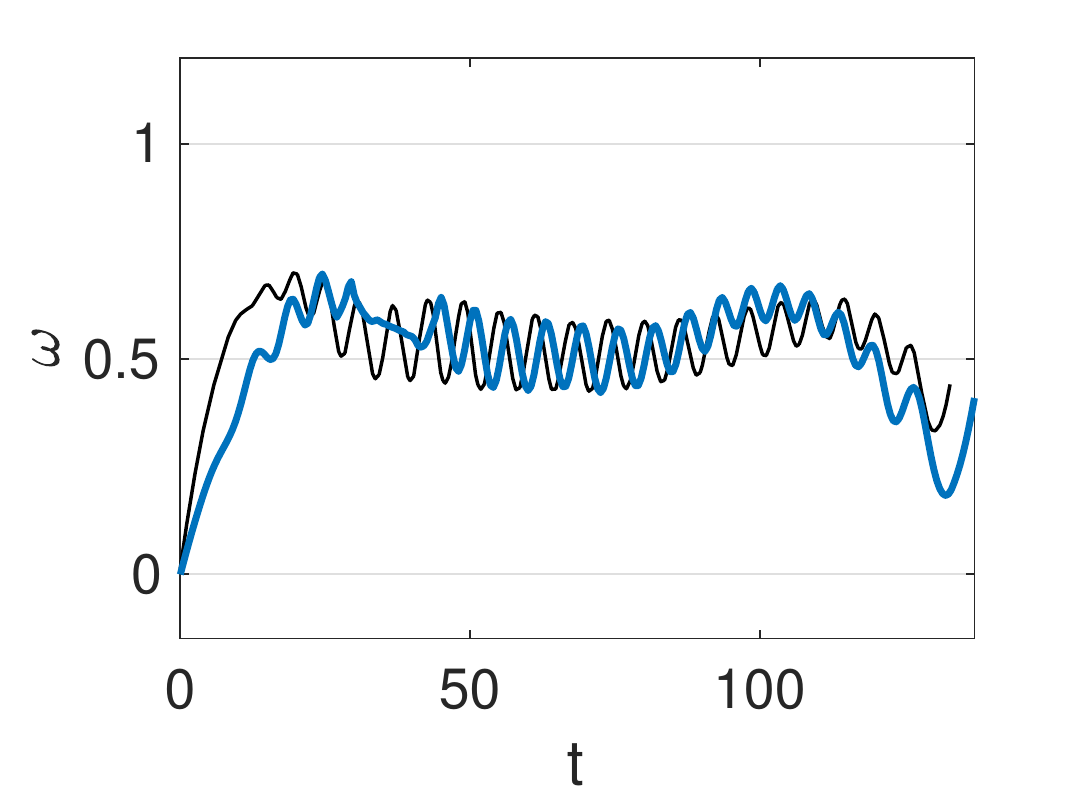}    
    \subcaption{} \label{fig:eno}\end{subfigure} \hfill \begin{subfigure}{0.325\textwidth} 
	\includegraphics[width=\textwidth, trim={8 3 20 10}, clip]{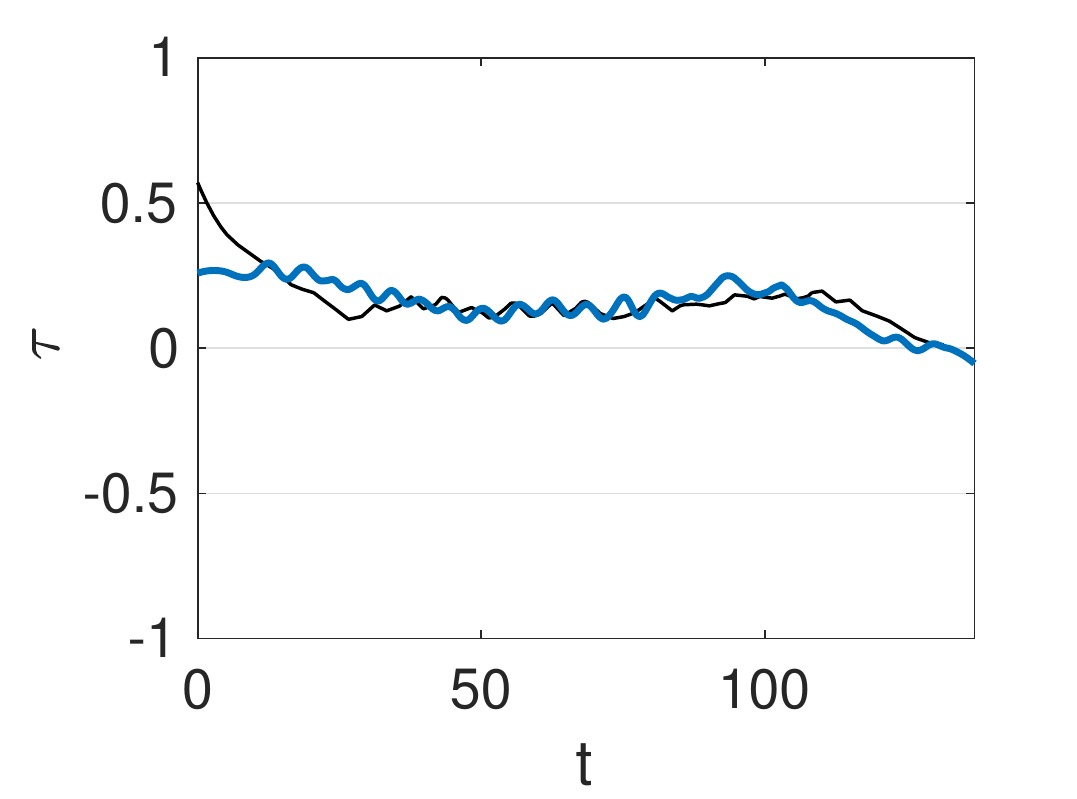}     
    \subcaption{} \label{fig:ent}\end{subfigure}
\caption{Energy-optimal (\subref{fig:enp})~$x$-$y$ trajectory, (\subref{fig:eno})~angular velocity, and (\subref{fig:ent})~control torque of the present gliding model obtained by reinforcement learning (blue lines) and optimal control by~\citep{paoletti2011} (black lines) for $\beta=0.1$ and $\rho^*=200$. Optimal control reaches the target position with energy-cost $E{=}4.4$ and time-cost $T{=}131$; RL achieves $T{=}137$, $E{=}4.3$, and lands with perching angle $44.6^\circ$.}\label{fig:enopt} 
\end{figure} 

\begin{figure} \centering 
    \begin{subfigure}{0.325\textwidth} 
	\includegraphics[width=\textwidth, trim={8 3 20 10}, clip]{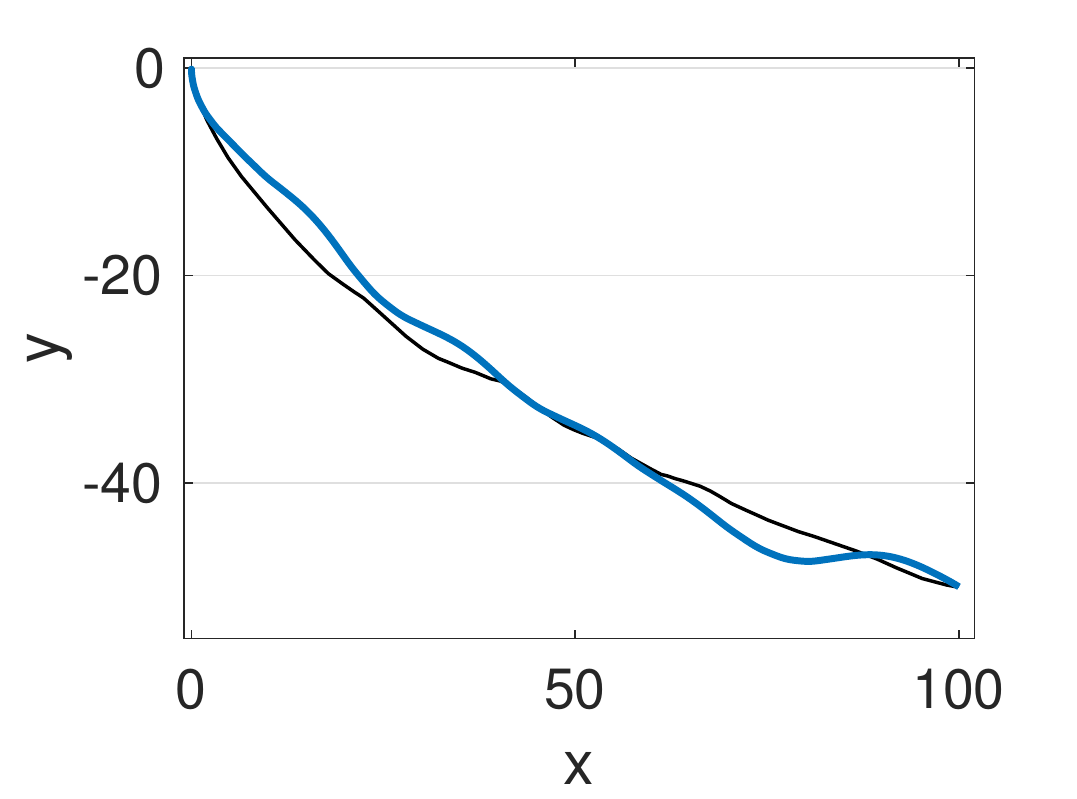}
	\subcaption{} \label{fig:tmp}\end{subfigure} \hfill \begin{subfigure}{0.325\textwidth} 
	\includegraphics[width=\textwidth, trim={8 3 20 10}, clip]{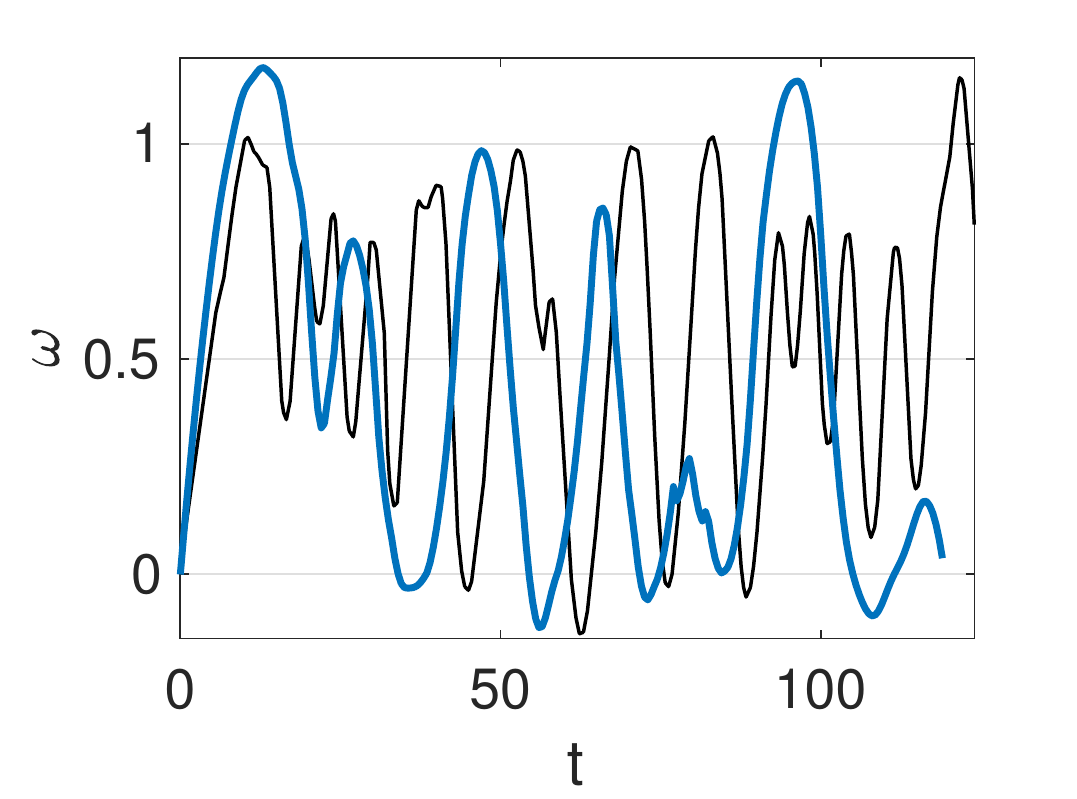}    
    \subcaption{} \label{fig:tmo}\end{subfigure} \hfill \begin{subfigure}{0.325\textwidth} 
	\includegraphics[width=\textwidth, trim={8 3 20 10}, clip]{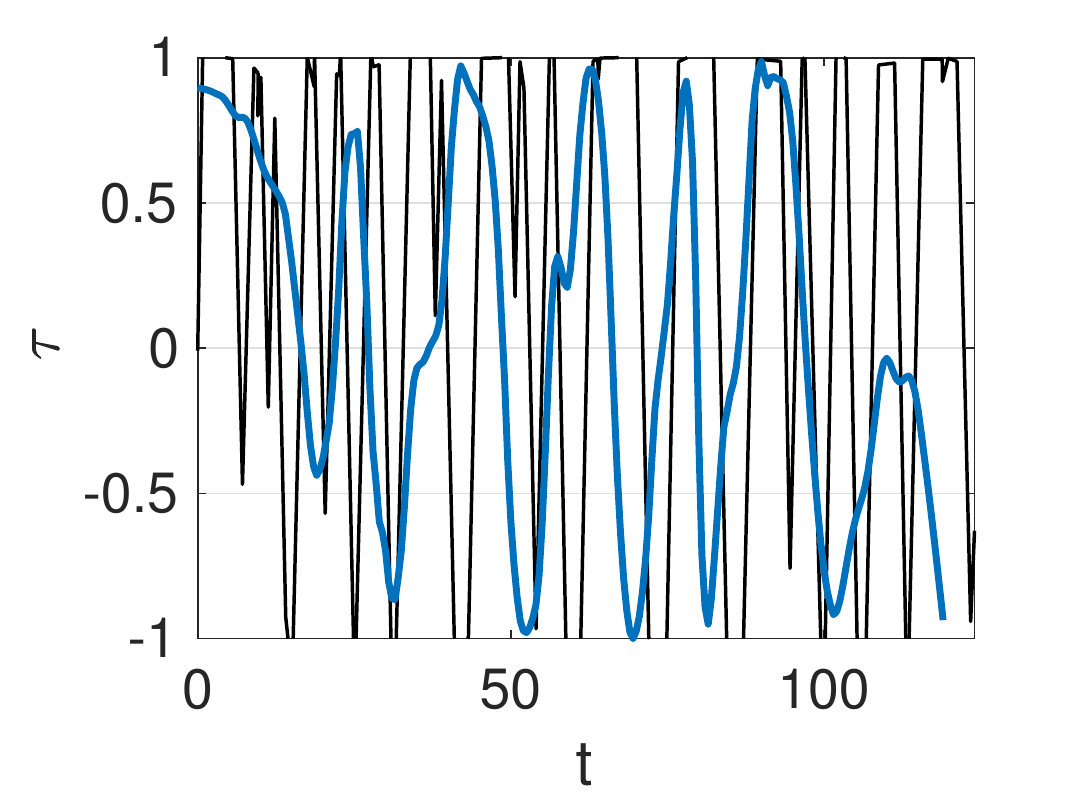}     
    \subcaption{} \label{fig:tmt}\end{subfigure}
	\caption{Time-optimal (\subref{fig:tmp})~$x$-$y$ trajectory, (\subref{fig:tmo})~angular velocity, and (\subref{fig:tmt})~control torque of the present gliding model obtained by reinforcement learning (blue lines) and optimal control by~\citep{paoletti2011} (black lines) for $\beta=0.1$ and $\rho^*=200$. Optimal control reaches the target position with time-cost $T{=}124$ and energy-cost $E{=}117$; RL achieves $T{=}119$, $E{=}48.7$, and lands with perching angle $44.9^\circ$.}\label{fig:tmopt} 
\end{figure}

Having obtained approximately-optimal policies with RL, we now compare them with the trajectories derived from optimal control (OC) by~\citep{paoletti2011} for $\rho^*=200$ and $\beta=0.1$. In figure~\ref{fig:enopt}, we show the energy optimal trajectories, and in figure~\ref{fig:tmopt} we show the time optimal trajectories. In both cases, we find that the RL agent surpasses the performance of the OC solution: the final energy-cost is approximately 2\% lower for RL and the time-cost is 4\% than that of OC. While in principle OC should find locally optimal trajectories, OC solvers (in this case GPOPS, see~\cite{paoletti2011}) convert the problem into a set of finite-dimensional sub-problems by discretizing the time. Therefore the (locally) optimal trajectory is found only up to a finite precision, in some cases allowing RL, which employs a different time-discretization, to achieve better performance. 

The RL and OC solutions qualitatively find the same control strategy. The energy-optimal trajectories consist in finding a constant minimal torque that generates enough lift to reach $x_G$ by steady tumbling flight. The time-optimal controller follows a "bang-bang" pattern that alternately reaches the two bounds of the action space as the glider switches between gliding and tumbling flight. 
However, the main drawback of RL is having only the reward signal to nudge the system towards satisfying the constraints. We can impose arbitrary initial conditions and bounds to the action space (Sec.~\ref{sec:rl}), but we cannot directly control the terminal state of the glider. Only through expert shaping of the reward function, as outlined in section~\ref{sec:rew}, we can train policies that reliably land at $x_G$ (within tolerance $\Delta x_T \approx 0.01$) with perching angle $\theta_G$ ($\Delta \theta_T \approx 0.5^\circ$).

One of the advantages of RL relative to optimal control, beside not requiring a precise model of the environment, is that RL learns closed-loop control strategies. While OC has to compute \textit{de-novo} an open-loop policy after any perturbation that drives the system away from the planned path, the RL agent selects action contextually and robustly based on the current state. This suggests that RL policies from simplified, inexpensive models can be  transferred to related more accurate simulations~\cite{verma2018} or robotic experiments (for example, see~\cite{geng2016}).

\begin{figure} \centering 
\begin{subfigure}{0.325\textwidth} 
	\includegraphics[width=\textwidth, trim={6 0 8 10}, clip]{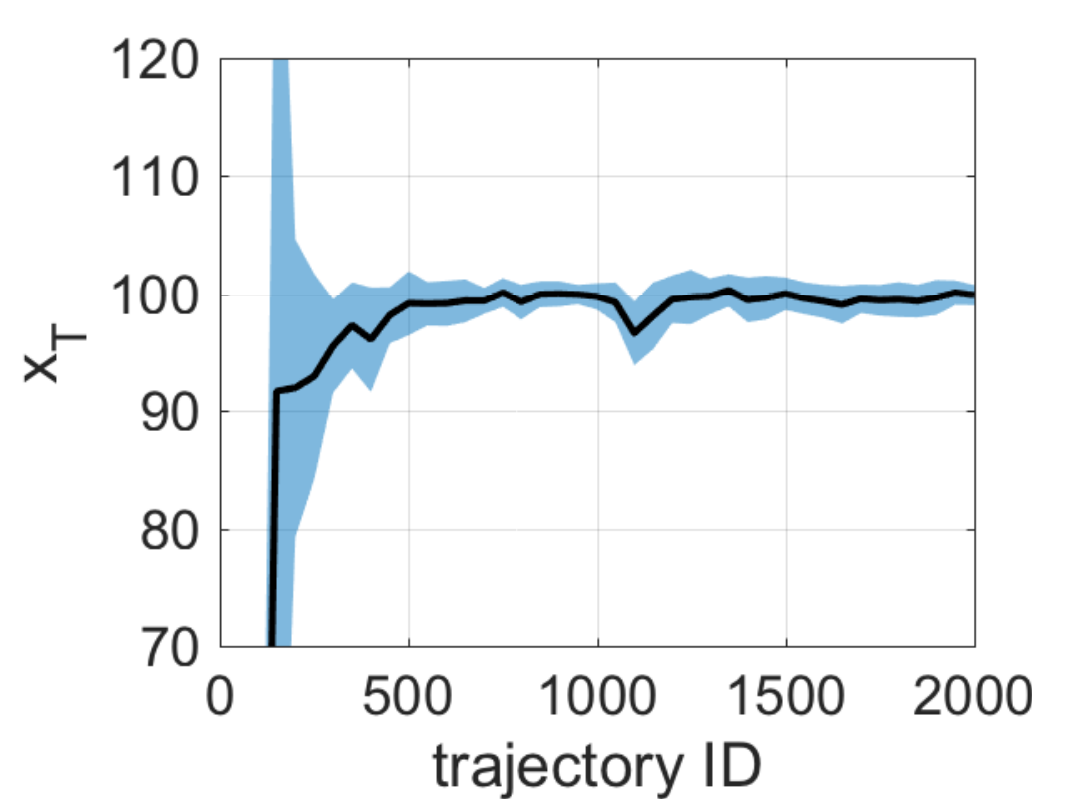}
	\subcaption{} \label{fig:algorace}\end{subfigure} \hfill \begin{subfigure}{0.325\textwidth} 
	\includegraphics[width=\textwidth, trim={6 0 8 10}, clip]{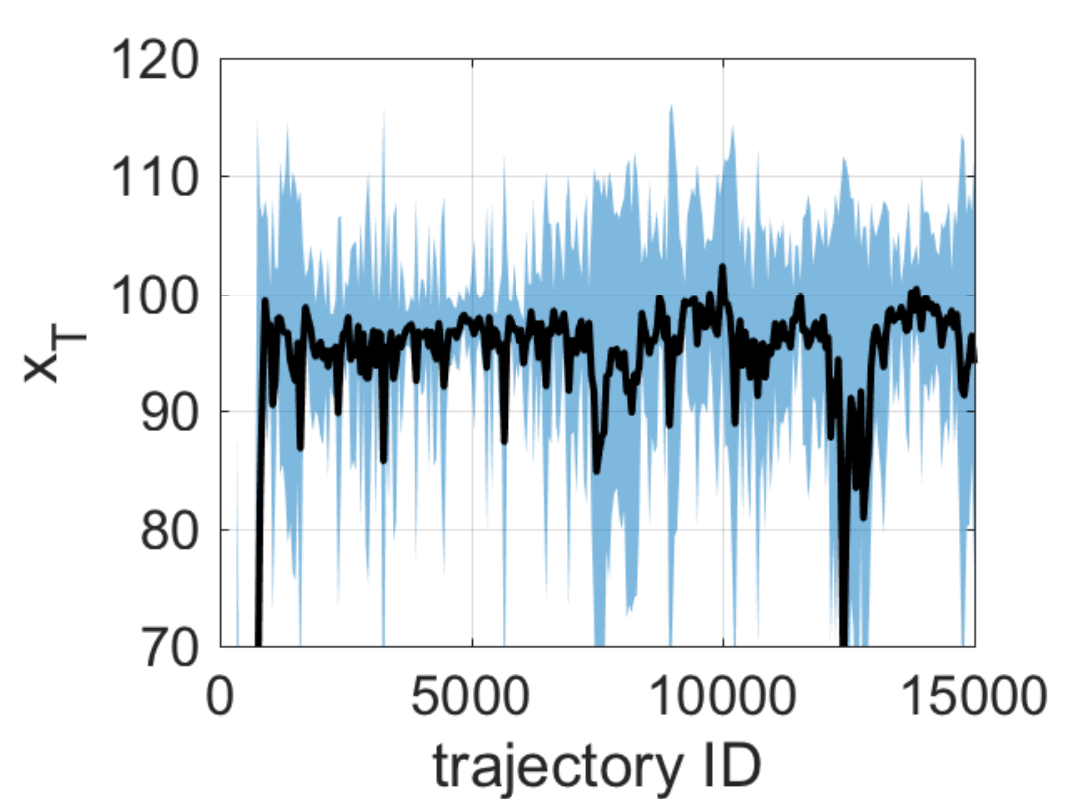}    
    \subcaption{} \label{fig:algonaf}\end{subfigure} \hfill \begin{subfigure}{0.325\textwidth} 
	\includegraphics[width=\textwidth, trim={6 0 8 10}, clip]{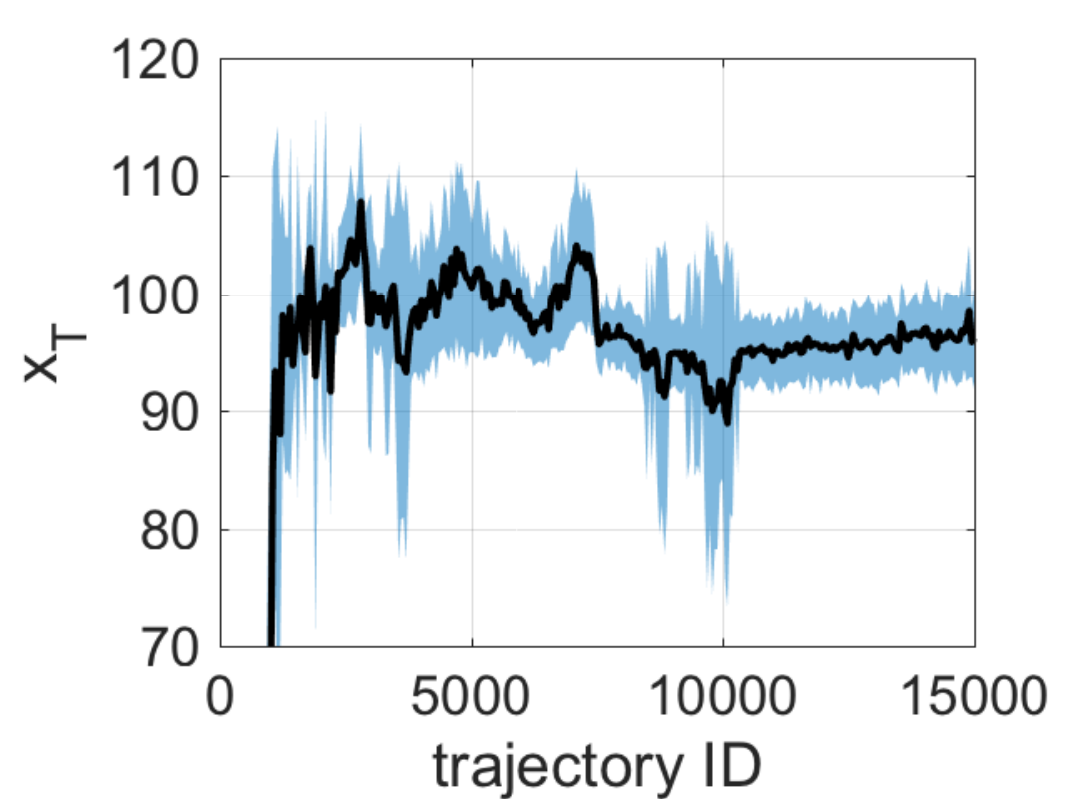}    
    \subcaption{} \label{fig:algoppo}\end{subfigure} 
	\caption{Distribution of landing positions during training for three RL algorithms: (\subref{fig:algorace})~Racer~\cite{novati2018} as described in section~\ref{sec:algo}, (\subref{fig:algonaf})~Normalized Advantage Functions~\cite{gu2016}, and (\subref{fig:algoppo})~Proximal Policy Optimization~\cite{schulman2017}. Racer reliably learns to land at the target position by carefully managing the pace of the policy update.}\label{fig:trainalgo}
\end{figure}

\section{Comparison of learning algorithms}\label{sec:comp}
The RL agent starts the training by performing actions haphazardly, due to the control policy which is initialized with random small weights being weakly affected by the state in which the agent finds itself. Since the desired landing location is encoded in the reward, the agent's trajectories gradually shift towards landing closer to $x_G$. 

In order to have a fair comparison with the trajectories obtained through optimal control, the RL agents should be able to precisely and reliably land at the prescribed target position. In general, the behaviors learned through RL are appealing, however, depending on the problem, it can be hard to obtain quantitatively precise control policies. This issue may be observed in figure~\ref{fig:trainalgo} where we show the time evolution of the distribution of terminal $x$-coordinates during training of three state-of-the-art RL algorithms. The {\tt Racer}  manages to reliably land in the proximity of $x_G$, after the first 1000 observed trajectories, with a mean squared error on the order of one. The precision of the distribution of landing locations, obtained here by sampling the stochastic policy during training, can be increased when evaluating a trained policy by choosing deterministically at every turn the action corresponding to its mean $m(s)$.

Normalized Advantage Function algorithm (NAF~\cite{gu2016}) is an off-policy value-iteration algorithm which learns a quadratic parameterization of the action value $Q^\theta(s,a)$, similar to the one defined in equation~\ref{eq:qq}. One of the main differences with respect to {\tt Racer} is that the mean $m^\theta(s)$ of the policy is not trained with the policy gradient (Eq.~\ref{eq:gradPi}) but with the critic gradient (Eq.~\ref{eq:gradQ}). While the accuracy of the parameterized $Q^\theta(s,a)$ might increase during training, $m^\theta(s)$ does not necessarily correspond to better action, leading  to the erratic distribution of landing positions in figure~\ref{fig:trainalgo}. 

Proximal Policy Optimization (PPO~\cite{schulman2017}) is an on-policy actor-critic algorithm. This algorithm's main difference with respect to Racer is that only the most recent (on-policy) trajectories are used to update the policy. This allows estimating $Q^\pi(s,a)$ directly from on-policy rewards~\cite{schulman2015b} rather than with an off-policy estimator (here we used Retrace~\ref{eq:retr}), and it bypasses the need for learning a parametric $Q^\theta(s,a)$. While PPO has led to many state-of-the-art results in benchmark test cases, here it does not succeed to center the distribution of landing positions around $x_G$. This could be attributed to an unfavorable formulation of the reward, or to the high variance of the on-policy estimator for $Q^\pi(s,a)$.

\section{Conclusion}\label{sec:end}
We have demonstrated that  Reinforcement Learning can be used to develop gliding agents that execute complex and precise control patterns using a simple model of the controlled gravity-driven descent of an elliptical object. We show that RL agents  learn a variety of optimal flight patterns and perching maneuvers that minimize either time-to-target or energy cost. The RL agents were able to match and even surpass the performance of trajectories found through Optimal Control. We also show that the the RL agents can generalize their behavior, allowing them to select adequate actions even after perturbing the system. Finally, we examined the effects of the ellipse's density and aspect ratio to find that the optimal policies lead to either bounding flight or tumbling flight. Bounding flight is characterized as alternating phases of gliding with a small angle of attack and rapid rotation to generate lift. Tumbling flight is characterized by continual rotation, propelled by a minimal almost constant torque. Ongoing work aims to extend the present algorithms to three dimensional Direct Numerical Simulations of gliders.
\subsubsection*{Acknowledgments}
We thank Siddhartha Verma for helpful discussions and feedback on this manuscript. This work was supported by European Research Council Advanced Investigator Award 341117. Computational resources were provided by Swiss National Supercomputing Centre (CSCS) Project s658.
\eject
\bibliography{glider}
\bibliographystyle{jfm}

\end{document}